\documentclass[a4paper, 10pt, conference]{ieeeconf}      

\IEEEoverridecommandlockouts                              

\usepackage{graphicx}
\usepackage{cite}
\usepackage{hyperref}
\usepackage{xcolor}
 
\author{Aditya Agarwal$^{*\dagger}$, Bipasha Sen$^{*\dagger}$, Shankara Narayanan V.$^{*\dagger}$, Vishal Reddy Mandadi$^{*\dagger}$, \\
Brojeshwar Bhowmick$^\ddagger$, K. Madhava Krishna$^\dagger$}

\usepackage[font=small]{caption}

\pdfoutput=1

\title{Approaches and Challenges in Robotic Perception for Table-top Rearrangement and Planning}
\begin{document}

\twocolumn[{%
\renewcommand\twocolumn[1][]{#1}%
\maketitle
\begin{center}
    \centering
    \captionsetup{type=figure}
    \includegraphics[width=\textwidth]{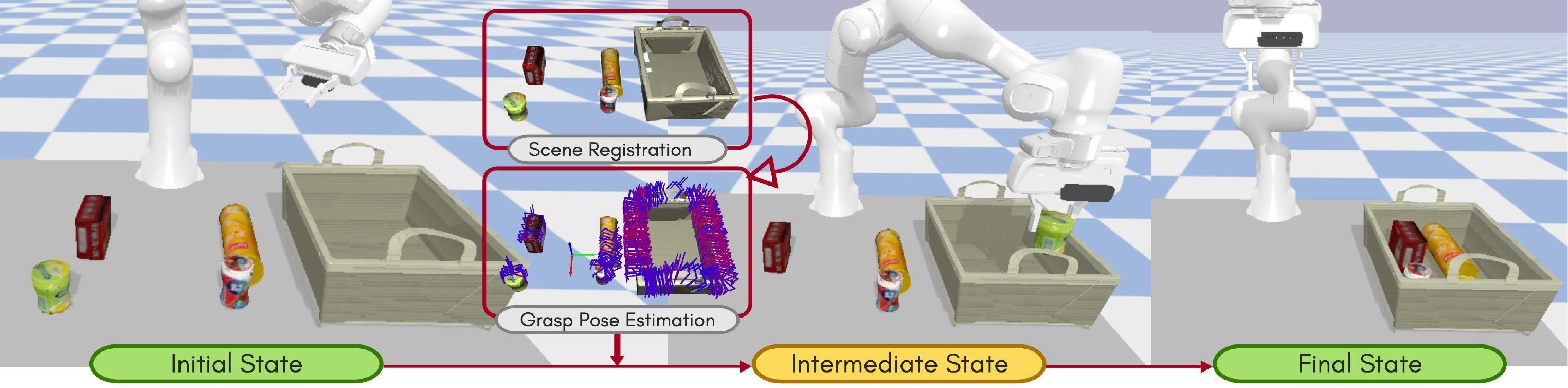}
    \captionof{figure}{Table-top rearrangement and planning involves arranging different objects on a table (left) to a given final configuration (right)}
    \label{fig:teaser}
\end{center}%
}]



\thispagestyle{empty}
\pagestyle{empty}

\begin{abstract}
Table-top Rearrangement and Planning is a challenging problem that relies heavily on an excellent perception stack. The perception stack involves observing and registering the 3D scene on the table, detecting what objects are on the table, and how to manipulate them. Consequently, it greatly influences the system's task-planning and motion-planning stacks that follow. We present a comprehensive overview and discuss different challenges associated with the perception module. This work is a result of our extensive involvement in the ICRA 2022 Open Cloud Robot Table Organization Challenge, in which we stood third in the final rankings. 


\end{abstract}
 %
\def\thefootnote{*}\footnotetext{Equal contribution (listed in alphabetical order)}
\def\thefootnote{$\dagger$}\footnotetext{Robotics Research Center, IIIT Hyderabad, India}
\def\thefootnote{$\ddagger$}\footnotetext{TCS Research, India}
\def\thefootnote{\arabic{footnote}}

\section{INTRODUCTION}
Table-top Rearrangement and Planning (TTRP) involves arranging a table-top scene given the final configuration of the scene (see Fig.~\ref{fig:teaser}). The final configuration is a list of objects and their final 6D pose on the table. For TTRP, one needs to undertake the following steps - (1) Register the 3D scene on the table, (2) Identify the poses of the objects in the scene, (3) Identify manipulation points (grasp points, push points, or suction points) for object manipulation, (4) Devise an order for arranging the objects from the identified initial poses to the given final poses, and (5) Plan the motion for each of the intermediate steps in (4).

In this work, we focus on the first three steps that make up the perception stack. We draw our observations from our involvement in the ICRA 2022 Open Cloud Robot Table Organization Challenge\footnote{\textcolor{blue}{\href{http://www.ocrtoc.org/}{http://www.ocrtoc.org/}}}\footnote{\href{https://github.com/OCRTOC/OCRTOC\_software\_package}{OCRTOC Github Baseline Package}}~\cite{ocrtoc} (OCRTOC). OCRTOC presents various interesting daily-needs use-cases for TTRP. Its setup includes a simulated table-top environment and a Franka Emika Panda robot arm with a two-finger gripper for manipulation\footnote{\href{https://github.com/justagist/franka\_ros\_interface}{https://github.com/justagist/franka\_ros\_interface}}. The objects in the competition are unknown; that is, at the time of training, the objects exposed are only 30\% of all the objects in the test scenes. Table-top items have huge variations across different homes; thereby constraining the items to known objects would not make the solution usable in real-life. The competition aims to devise a generalized approach that can solve various situations from as simple as -- placing the objects on the table into a basket -- to -- arranging huge books by performing actions like swapping and stacking. The variations in the type of tasks and the objects involved make every step of the pipeline challenging. 

In the following sections, we discuss each step (of the perception stack) in detail, followed by the challenges encountered in each step during our near-year-long involvement in the challenge. We also provide possible solutions and future directions of research. Although we draw inspiration from the OCRTOC setup, we assume settings beyond the OCRTOC environment while proposing potential research areas.
We hope to encourage research in the space of TTRP (especially on the challenging perception stack) and hope that TTRP turns from a task in research labs to a real-life daily-used home and industry soon. 

\section{Perception in Typical Table-top Rearrangement and Planning Pipelines}
A typical TTRP pipeline can be divided into (1) Perception and (2) Planning stack. The perception stack contains three significant aspects - 1. scene registration, 2. (unknown) object detection, and 3. detecting manipulation (grasp/push) points. 

\textbf{Scene Registration: }Albeit one of the most common steps in robotics, scene registration presents exciting challenges in TTRP. This stage involves capturing the 3D scene from different poses, then registering the local scenes into one global scene. The poses are predefined locations that maximize the view at each location to make the final registration dense. The registration is finetuned using ICP to create a dense pointcloud of the table-top scene. This step is crucial as it determines the manipulation points (grasp poses, push points, suction points) for the objects in the scene. 

\textbf{Object Detection and 6D Pose Estimation: }In this step, we determine the objects on the scene along with their 6D (translation $+$ orientation) pose. There has been extensive research on this topic, specifically known object instance~\cite{posecnn, pvnet, pix2pose, cosypose, deepim} and known object category~\cite{nocs, irshad2022centersnap, categorylevel, fsnet} pose estimation using 2D images~\cite{localsurfacepose, posecnn, pvnet, pix2pose, dscposenet} or 3D scenes~\cite{stablepose, e2ecad, ffb6d, implicit3d, pvn3d}. In known object instances, we observe the exact object instance at training. In known object \textit{category}, the exact object category instead of an instance is observed. A category can be a camera. Here, the instances would be Canon GTX II or a Nikon D780. Some of the most commonly used datasets for this step are - YCB Video~\cite{posecnn}, NOCS~\cite{nocs}, and T-LESS~\cite{tless}. Even though the progress on known-instance / category object pose estimation has been extensive, there has been little progress on unknown-category object-pose estimation. It is an extremely challenging task due to the amount of variation in the type of objects, ranging from a simple pen, soap, and bottles to more complicated objects like laptops, bobbleheads, or even deformable objects like clothes and chargers.

\textbf{Manipulation Points: }There are several different ways an arm could manipulate an object: Push, Grasp, and Suction. 
Each of these modes has its classic use case (such as pushing when an object is not small enough for grasping) and needs careful selection of the corresponding manipulation points on the objects. The two fundamental challenges in this step are: (1) deciding the manipulation mechanism (grasping, pushing, or suction) and (2) deciding the point of contact (manipulation point) given the object and the scene. In this work, we only discuss grasping as a manipulation approach. There have been many recent advances in grasp proposal approaches, including model-free (unknown-category object) grasp proposals~\cite{6dof, contact,1billion,realtimegrasping}. GraspNet-1Billion~\cite{1billion}, trained on 1 billion grasp poses, proposes many grasp poses given an object pointcloud. It, however, does not extend well to irregular and large scene pointcloud. Contact-GraspNet~\cite{contact} proposes grasps directly on an entire scene. Although many robust grasp proposal networks exist, selecting the best grasp from the generated proposals for manipulating a given object in a scene is an important area that has been less explored~\cite{suitablegrasping} needing the presence of tactile sensors.

\begin{figure}
    \centering
    \includegraphics[width=\linewidth]{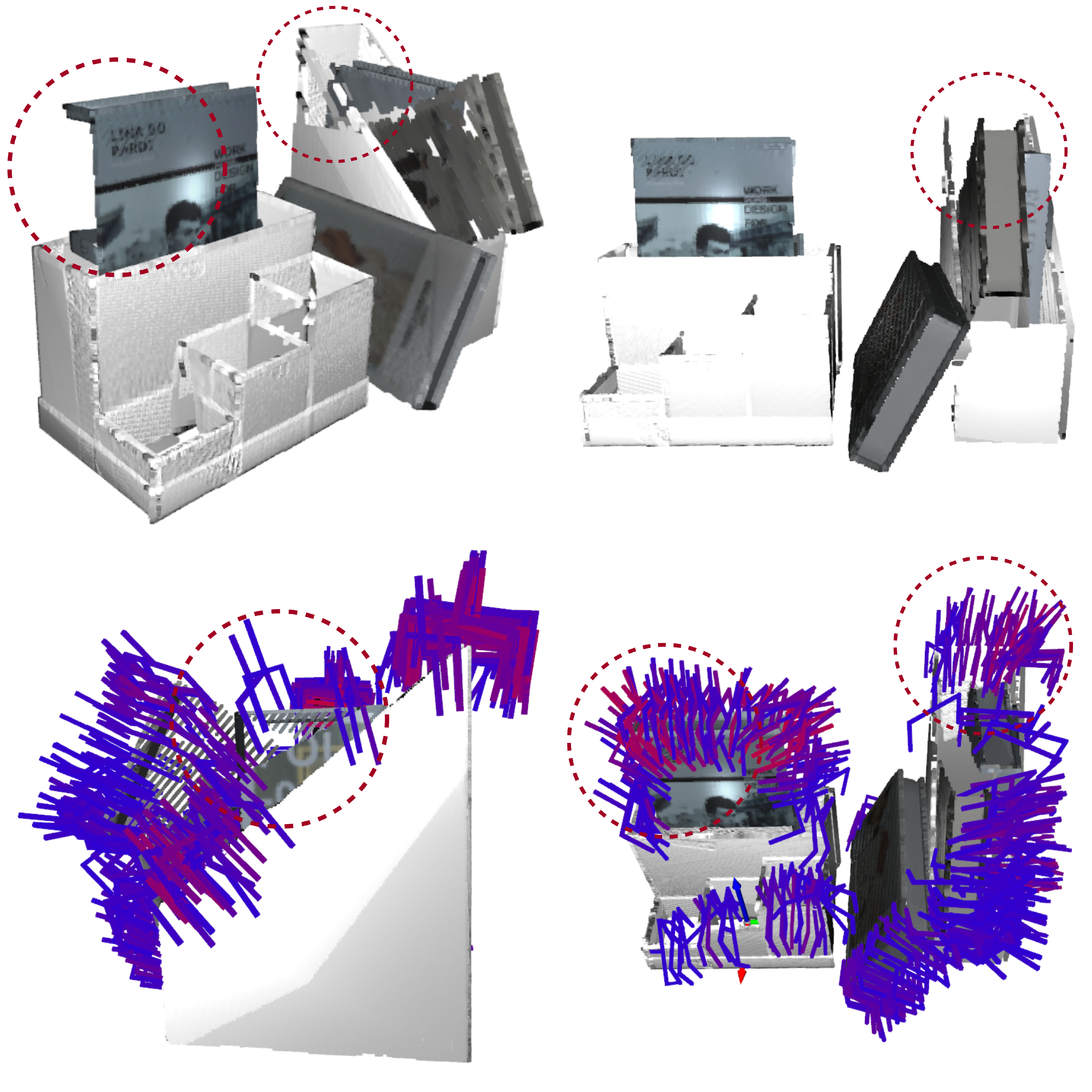} 
    \caption{An example of a noisy table-top reconstruction with holes (red cirles, top-half) leading to bad grasp proposals in the regions with holes (red cirles, bottom-half).}
    \label{fig:holes}
\end{figure}

\section{Challenges and Discussion}

In this section, we provide a comprehensive list of the critical challenges in TTRP. We formulate simple yet effective strategies for tackling some of these challenges by adapting existing state-of-the-art methods. We propose potential research areas for the other challenges that simple techniques cannot solve. The observations in this section are made on Pybullet\footnote{\textcolor{blue}{\href{https://pybullet.org/wordpress/}{https://pybullet.org/wordpress/}}} and Sapien\footnote{\textcolor{blue}{\href{https://sapien.ucsd.edu/}{https://sapien.ucsd.edu/}}}. However, the challenges and approaches proposed here extend to any other simulation or real environment.

\subsection{Scene registration}

Scene registration is a fairly simple step that involves capturing the scene pointcloud from different camera poses, followed by scene registration using the known poses and ICP. It is critical for (1) running grasp proposal networks~\cite{6dof,contact,1billion,realtimegrasping} and (2) object-pose estimation by registering the object's 3D representation on the scene pointcloud. However, this step suffers from two fundamental challenges. Firstly, capturing the images from several locations consumes significant time (an average of 1.5 minutes for three camera poses). Secondly, it is common to end up with a noisy registration made of many critical holes. Capturing a local scene from carefully calibrated viewpoints leads to an overall dense and detailed scene representation. Even then, it may be challenging to ensure the coverage of each point location in the 3D table-top volumetric scene (given the variations in the object and scene structure), leading to holes in the registered scene. These holes consequently lead to faulty grasp proposals in these regions (see Fig.~\ref{fig:holes}), which subsequently lead to collisions.


\textbf{Noisy scene registration: }ICP or deep-learning-based approaches~\cite{e2ecad} can turn a noisy or sparse scene into a dense and detailed scene by registering the objects' pointclouds in the scene pointcloud. This way, one can even decrease the number of camera poses for registration to obtain sparse registration and then use the mentioned approaches to recreate the dense scene. Despite being fairly intuitive, this approach faces the following drawbacks: (1) Additional time overhead - starting with a random initial object position for ICP based registration will be costly. It could take up to several minutes to register all the objects in the scene. Moreover, this step would be a complete misspend for smaller objects easily lost in the scene point cloud. Another way is initializing the objects' pose with the near-perfect approximation of the ground-truth intial 6D pose (estimated by performing object-pose estimation) followed by the usual ICP step for registration. A good initialization will make the time-overhead negligible. This would be dependent on an excellent performance by the object-pose detection algorithm. However, object-pose estimation is not always accurate, especially for scenes with unknown-category objects. (2) Dependency on a 3D representation of the object - As the number of objects in TTRP increases (specifically, to entertain real-life use-cases as it is seldom the case that two different tables will have \textit{any} overlapping objects), the availability of the 3D object representation becomes less likely. This makes the dependency on a 3D object representation a bottleneck for scalability and generalization. 
\underline{Potential Research Direction}: Given a noisy scene, we can exploit a geometrical prior such as surface normals / continuous surfaces to complete the pointcloud. Such an approach could be independent of specific object models. Instead, an underlying geometrical consistency of the overall scene can be exploited to find and interpolate holes between neighboring points.


\textbf{Dealing with time overhead: }Capturing the scene from a single view followed by dense registration using ICP or \cite{e2ecad} as mentioned above can significantly reduce the time taken. However, the scene pointcloud would be noisy, especially as many objects would be half or entirely invisible due to occlusion. Another way can be to avoid the scene registration step entirely. To understand this, we need to look at why we need scene registration at all? The fundamental use of this step is to obtain grasp proposals of the objects on the scene~\cite{6dof, contact,1billion,realtimegrasping}. Instead of depending on the entire scene pointcloud, the grasp proposals can be made using the pointcloud of the individual objects. First, using GraspNet-1Billion~\cite{1billion} (trained on individual objects pointclouds) we can get grasp proposals on the object. Next, we can rely on the object-pose estimation step to determine the object's location in the scene. We then transfer the grasp proposals to the scene using the relative transformation between the object's pose in the canonical and scene forms. However, a bottleneck in terms of scalability and generalization arises here as we depend on the 3D object representation. Also, a dependency on accurate pose estimation is founded. \underline{Potential Research Direction}: Can we predict the grasp poses for the objects in the scene using only a 2D RGB scene image? How can we exploit the geometrical priors and structural information in 2D to determine the grasp poses in 3D? Today, approaches are dependent on a 3D geometrical structure for grasp proposals. Estimating robust grasps from a single RGB / even an RGB-D image can have huge impacts on several related challenges -- such as avoiding grasp poses in regions with holes.


\subsection{6D Object Pose Estimation}
\label{section:objectpose}
Given the target 6D pose of an object, we need to estimate the initial 6D pose of the object in the scene. Object pose estimation can be broadly categorized into three categories: (1) Known Instance, (2) Known Category and (3) Unknown Category. Although (1) and (2) have witnessed extensive research, there has been little headway in (3). 

An indirect approach can be to use classical~\cite{sift} or deep-learning~\cite{superglue} based feature matching algorithms to match scene images with the object projections in 2D. We start with the RGB scene captured from different known poses and the 2D projection of the 3D object representation to different known poses. Next, we find a scene and the corresponding object projection such that they have the maximum feature matches. As both the poses (object and scene) are known, we can now solve for the object's exact global 6D pose. This approach is very accurate in cases where the objects have distinct shapes and structures. However, feature matching operates in grayscale; thus, this approach falters on similar-shaped objects (such as strawberries/oranges/apples). Also, the entire process consumes a significant amount of time for accurate pose-estimation (an average of 5 minutes for all the objects in the scene on modern GPUs), making it difficult to be adopted in real-world scenarios with real-time runtime.  

A second indirect approach uses ICP to register the object pointclouds in the scene pointcloud. However, this step is slower than feature-matching and leads to comparable pose-estimation performance making this approach less desirable.

\begin{figure}
    \centering
    \includegraphics[width=\linewidth]{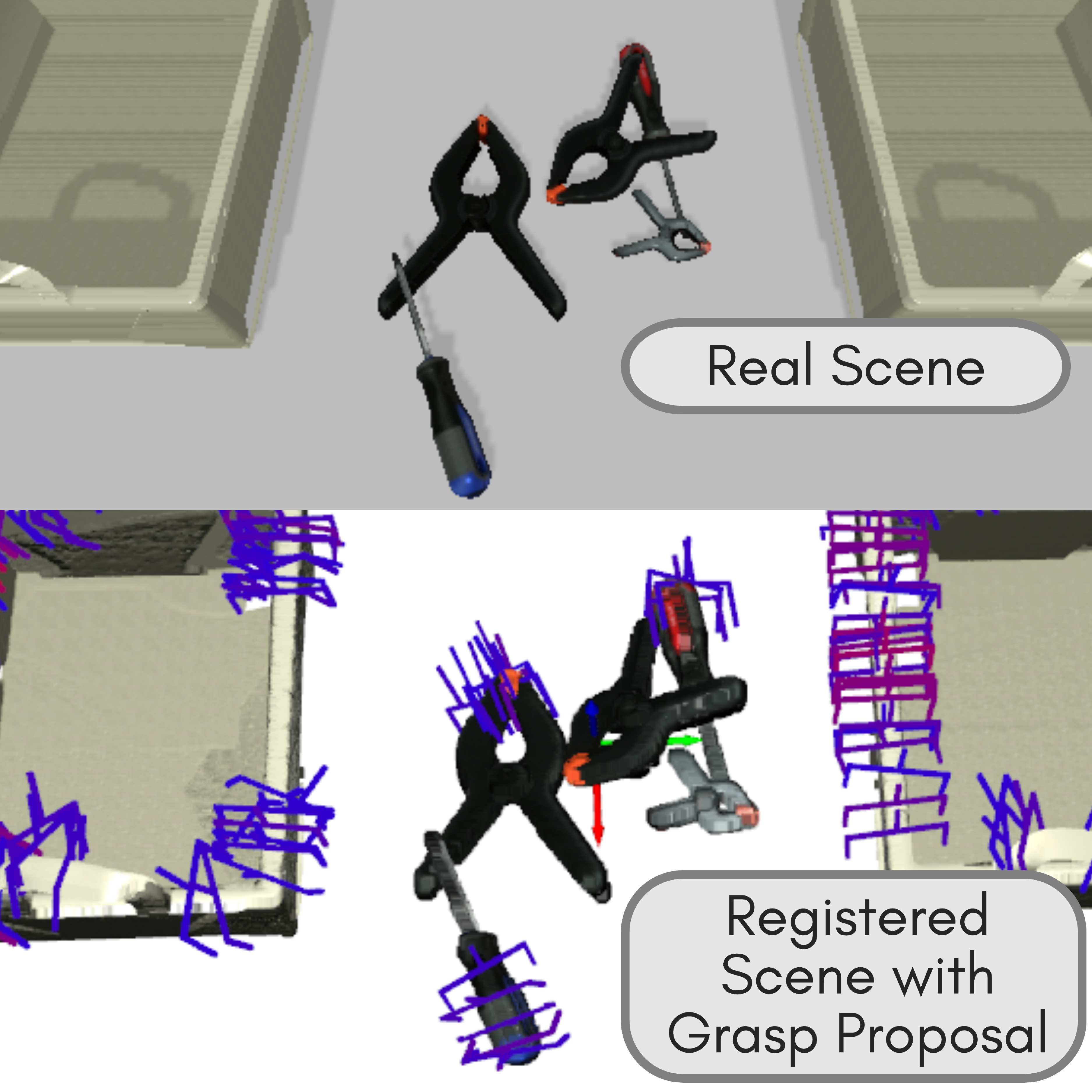} 
    \caption{An example of cluttered environment. Note that the top black plier and the smaller gray plier do not have any grasp proposals.}
    \label{fig:cluttered}
\end{figure}

Both of the approaches mentioned above depend on the existence of a 3D object representation or at least the objects' multi-view 2D projection. They also do not work well on smaller objects. The feature matching-based approaches do not generalize to large objects either (such as books in Fig.~\ref{fig:holes}). Another area where they struggle is scenes with clutter (see Fig.~\ref{fig:cluttered}). 
\underline{Potential Research Direction}: We can rely on robust finetuned object descriptions obtained from large architectures like Resnet~\cite{resnet}, VGGNet~\cite{vggnet}, 
pretrained on massive datasets like ImageNet~\cite{imagenet}. First, assume a single-view/multi-view 2D projection of the 3D object representation. Next, use a pretrained/finetuned segmentation network such as Mask R-CNN~\cite{maskrcnn} to detect and segment the objects in the scene. Here, the task reduces to matching the segmented objects' descriptors with the descriptor of the objects' projections. The descriptors are obtained from the large networks mentioned above. In the case of single-view projection, the pose difference between the projection and the segmented object would need to be estimated. For multi-view, the projections' pose that has the highest match with the segmented objects' pose in the scene can be used to regress the final 6D pose. The primary challenge here is to devise a mechanism that can reliably match the descriptors between the object projection and the segmented scene.

\subsection{Manipulation Points}

Contact-Graspnet~\cite{contact} takes an entire pointcloud as input and determines the grasp-points on the entire scene, inherently taking care of collision and clutter. For example, given two objects placed side-by-side. One of these objects may be graspable from the side attached to the other object. The attached side could get some proposals if the grasp proposals are made for objects individually. During the grasp selection, an explicit check would be needed to avoid the poses that are in collision with the neighboring object. In the case of Contact-Graspnet, which proposes grasps for the entire scene, the side-view of the object would not be visible, and thus there would be no proposals on that region. In this section, we will specifically focus on the grasp assignment logic. Grasp assignment logic concentrates on determining the best grasp for a selected object from a set of proposed grasps given the object's placement in the scene.

One naive way is to assign the grasp pose closest to the object's centroid. However, as can be seen in Fig.~\ref{fig:cluttered} (bottom), this is not a reliable approach as a grasp on a different object could be closer to the object in question. Another challenge is when an object is not assigned any grasp at all. For instance, in Fig.~\ref{fig:cluttered} (bottom), one of the pliers does not have any grasp proposals. If one naively selects the grasp closest to the centroid, the grasp on a different object (which might be far from the object in question) would be assigned to it. One way to counter this issue is placing a threshold on the distance between the object's centroid and the grasp pose. However, this is not object agnostic, as can be seen in the same figure; the threshold for the tray would be very different from the threshold on the small plier. 

The above issues (wrong grasp assignment) can be tackled by using the object's pointcloud outside of the scene. First use the pointcooud to obtain the grasp proposals through \cite{6dof,contact,1billion}. Next, transfer the proposals to the scene using the object's detected/given initial pose in the scene. Here, we would know exactly which grasp proposal is assigned to which object. However, these proposals would not inherently avoid collision as in the case of Contact Graspnet that makes the grasp proposal for the entire scene. 

A different challenge when dealing with grasp assignments is the object's structure. Since Contact-Graspnet considers the entire scene pointcloud, it does not propose a grasp that is structure-aware. For example, the best way to hold a plier would be on the fingers, whereas for a bottle, the best pose would be the one closest to the centroid. 

\underline{Potential Research Direction}: The existing grasp proposal algorithms try to determine any valid graspable points in the scene. It does not take into account the structural properties. Also, it does not know individual object properties (plier vs. a bottle). Moreover, the grasp poses are proposed unevenly across the objects. Despite the hundreds of proposals, only a few are meaningful, given the scene, the object, and the object's orientation. Thus instead of only focusing on meaningful grasps, one could concentrate on grasp ranking. This could be devised as an interesting reinforcement learning problem as this would need sufficient hit-and-trials before the arm figures out the best grasp for an object in the given scene configuration for the highest pick success. Some existing approaches do try to determine the best grasp pose given the property of an object~\cite{suitablegrasping}. However, such works are limited by the dependency on extra tactile sensors. Moreover, considering the scene characteristics for determining the best grasp pose is an exciting area to explore.

\underline{Object Verification}: Grasp assignment in cluttered environments is tough to handle. In fact, despite selecting the correct grasp, the gripper may pick up the wrong object due to its proximity to the correct object. To tackle this issue, we propose an additional check -- on-the-fly grasp verification that can verify if the object grabbed matches the intended object. The picked object can be accurately retrieved by performing scene segmentation at the position where the object is suspended from the gripper. This segmented object can then be matched with the intended object's projection and the other object's projection to determine if the correct object was picked. An approach similar to one described in subsection.~\ref{section:objectpose} can also be used for the verification.

\section{Conclusion}
We provide a comprehensive overview of the typical steps involved in the perception stack of a Table-top Rearrangement and Planning (TTRP) pipeline. We then discuss each step in detail, the challenges faced in each step, potential solutions and research areas. TTRP in an area of active research, despite that, it is largely unsolved. We hope to encourage research in this area and to also make the pipeline more generalizable, scalable, and faster.



\section{Acknowledgements}

We thank TCS Research for funding our research. We also thank Abhishek Chakraborty, Marichi Agarwal, Sourav Ghosh, and Dipanjan Das, who are research scientists at TCS research for their valuable guidance and help.

\bibliographystyle{IEEEtran}
\bibliography{main}

\end{document}